\documentclass[11pt]{article}

\usepackage[preprint]{acl}

\usepackage{times}
\usepackage{latexsym}

\usepackage[T1]{fontenc}

\usepackage[utf8]{inputenc}

\usepackage{microtype}

\usepackage{inconsolata}

\usepackage{graphicx}

\usepackage{enumitem}
\usepackage{amsmath}
\usepackage{algorithm}
\usepackage{algorithmic}
\usepackage{booktabs}
\usepackage{multirow}

\title{SpecPV: Improving Self-Speculative Decoding for Long-Context Generation via Partial Verification}

 \author{Zhendong Tan,\, Xingjun Zhang, \, Chaoyi Hu, \, Junjie Peng, \, Kun Xia\\
         School of Computer Science and Technology, Xi'an Jiaotong University 
         \\ \texttt{772316639@stu.xjtu.edu.cn, xjzhang@xjtu.edu.cn}}

\begin{document}
\maketitle
\begin{abstract}
Growing demands from tasks like code generation, deep reasoning, and long-document understanding have made long-context generation a crucial capability for large language models (LLMs). Speculative decoding is one of the most direct and effective approaches for accelerating generation. 
It follows a draft–verify paradigm, where a lightweight draft model proposes several candidate tokens and the target model verifies them.
However, we find that as the context length grows, verification stage gradually becomes the dominant bottleneck. 
To further accelerate speculative decoding in long-context scenarios, we introduce \textbf{SpecPV}, a self-speculative decoding approach that performs fast verification using partial key–value states (KV) and periodically applies full verification to eliminate accumulated errors.
We validate SpecPV across multiple long-context benchmarks and models, including LLaMA-3.1-8B-Instruct and Qwen3-series.
Experimental results show that SpecPV achieves up to 6× decoding speedup over standard autoregressive decoding with negligible degradation.
Our code is available at: \url{https://github.com/TanZhendong/SpecPV}.
\end{abstract}

\section{Introduction}
\label{sec:Introduction}
Long-context generation has become an essential capability for large language models (LLMs). Most modern models, such as DeepSeek and Qwen3 \cite{liu2024deepseek, yang2025qwen3}, now support context windows more than 128K tokens. 
The demand for long-context generation arises in two main scenarios. 
On the one hand, as LLMs become more powerful, applications such as document analysis, summarization, and code generation require processing much longer inputs, driving the need for extended context lengths. 
On the other hand, as think modes become more prevalent \cite{deepseekr1,openaio1}, models are required to explicitly generate their reasoning process. This often leads to longer and even verbose outputs, which further amplifies the challenge of efficient long-context generation.

Many approaches have been proposed to improve the efficiency of long-context model generation, such as sparse attention \cite{tang2024quest, streamingllm}, KV cache quantization \cite{liu2024kivi}, and even the exploration of new architectures \cite{wei2025deepseekocr, yang2024gated}. Beyond these, one of the most direct and effective methods is \emph{speculative decoding}. 
During autoregressive generation, the model can only predict one token at a time, requiring repeated forward passes when producing long sequences. Speculative decoding alleviates this limitation through a \emph{draft–verify} framework: a lightweight draft model first generates multiple candidate tokens, which are then verified by the target LLM in a single forward pass. Since the draft model is much faster and multiple tokens can be accepted at once, this approach substantially reduces decoding latency.

\begin{figure}[t]
\begin{center}
\centerline{\includegraphics[width=\columnwidth]{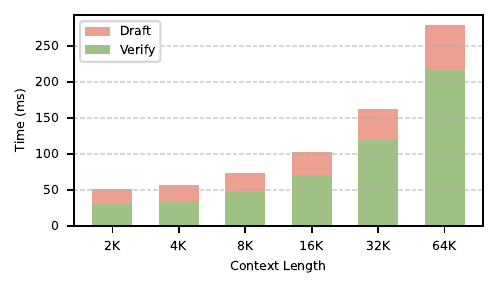}}
\caption{Drafting and verification time of EAGLE-3 speculative decoding on LLaMA-3.1-8B-Instruct. As the context length increases, verification gradually becomes the dominant bottleneck.}
\label{fig: draft_verify_time}
\vspace{-2.5em}
\end{center}
\end{figure}

\begin{figure*}[t]
\begin{center}
\centerline{\includegraphics[width=2\columnwidth]{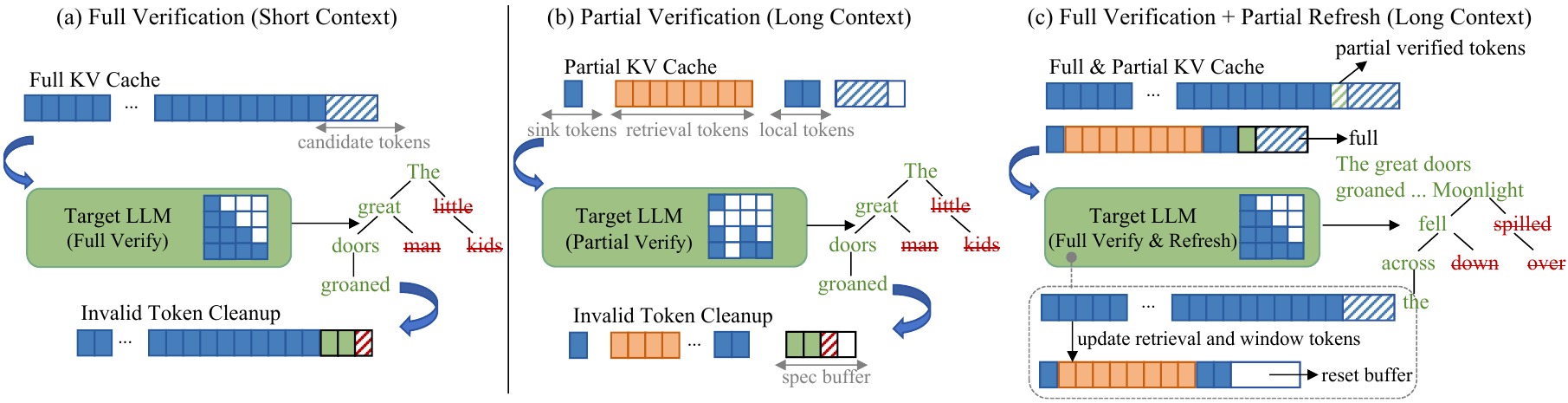}}
\vskip 0.1in
\caption{Illustration of three verification processes in SpecPV. For short context, we adopt classic full verification, whereas for long context, we use partial verification to improve efficiency. Periodic full verification eliminates accumulated errors and refreshes the partial KV cache. Taken together, these modes balance efficiency and accuracy across different context length.}
\label{fig: specpv framework}
\vspace{-2em}
\end{center}
\end{figure*} 

Recent studies have explored applying speculative decoding to long-context generation in pursuit of lossless acceleration \cite{suntriforce, wu2025tokenswift, sadhukhanmagicdec}. When handling long contexts, the primary challenges lie in the rapidly growing KV cache and the complexity of attention. Consequently, these methods typically focus on designing draft models with lightweight KV cache, while relying on full KV during verification to ensure lossless.

However, when applying EAGLE-3 \cite{li2025eagle} to long-context generation, we measure the average drafting and verification time, as shown in Figure \ref{fig: draft_verify_time}. For shorter contexts, although each draft step is relatively fast, searching and generating a tree-structured set of candidate tokens still incurs a non-negligible cost. As the context length increases, however, the verification time grows from around 60\% to nearly 80\%, gradually becoming the dominant bottleneck.
This observation raises a fundamental question: \emph{is full verification truly necessary in long-context speculative decoding, or can reliable verification be achieved using only partial KV states, thereby substantially reducing verification overhead?}

A natural concern arises that partial verification might harm model performance. However, numerous studies \cite{zhang2023h2o, tang2024quest, xuxattention} have shown that attention in LLMs exhibits strong sparsity, especially when processing long context.
This indicates that, by employing carefully designed retrieval strategies, one can achieve effective verification using partial KV cache with negligible accuracy loss.

Building on this insight, we propose a partial verification approach for self-speculative decoding. In this framework, the target LLM and the lightweight draft module alternately perform forward passes. During verification, the target LLM obtains the layer features of accepted tokens, which are reused to guide subsequent drafts. 
Considering that LLMs inevitably accumulate errors during partial cache generation, we replace only part of the verification steps with partial verification, while periodically performing full verification to rectify deviations and ensure that the generated results remain consistent with the original outputs.
Moreover, since the partial KV cache is much smaller than the full one, under memory-constrained settings we can offload only the full KV cache while keeping the partial cache on-device, thereby achieving even better speedup.
Finally, we evaluate the model performance and efficiency of SpecPV across different models and datasets using the EAGLE-3 draft \cite{li2025eagle}. Experimental results demonstrate that SpecPV incurs only negligible accuracy loss and provides substantial efficiency gains, achieving up to 6× decoding speedup compared to standard autoregressive decoding and around 2× speedup over full verification at a 60K context length.

In summary, our key contributions are as follows:
\begin{itemize}[itemsep=1pt, topsep=1pt, parsep=1pt]
    \item We summarize the framework of Self-Speculative Decoding and identify its bottlenecks in long-context generation.
    \item We propose SpecPV, a speculative decoding framework with partial verification, which further accelerates long-context generation.
    \item We evaluate SpecPV across different models and datasets, showing up to 6× decoding speedup with negligible accuracy loss.
\end{itemize}

\section{Related Work}

\paragraph{Speculative Decoding}
Speculative decoding is a widely used approach for losslessly accelerating autoregressive models. \citet{leviathan2023fast} introduced Speculative Sampling and demonstrated that it is theoretically equivalent to standard sampling.

A key challenge in speculative decoding is how to organize candidate tokens to improve the accepted length during drafting. Classical approaches employ a small speculative model to generate a single candidate sequence. SpecInfer \cite{miao2024specinfer} organizes candidate tokens into a tree, effectively increasing the number of accepted tokens by expanding the candidate space. Building on this, subsequent works \cite{li2024eagle2, liu2024kangaroo} have explored dynamic tree construction based on the draft model’s predicted distribution.

Another key challenge  lies in designing a fast yet effective draft model, as its speed and accept length directly determine the overall acceleration. Traditional approaches typically employ an independent small model, such as LLaMA-68M, to serve as the draft \cite{miao2024specinfer}.
Several studies leverage part of the target LLM’s layer features to construct the draft model, a paradigm commonly referred to as self-speculative decoding. Specifically, some studies implement this paradigm through layer skipping or early exiting \cite{zhang2024draft, elhoushi2024layerskip, liu2024kangaroo, xia2025swift}. In contrast, methods such as EAGLE \cite{li2024eagle}, Medusa \cite{cai2024medusa}, and Multi-Token Prediction (MTP) \cite{gloeckle2024better} directly reuse the top-layer features as part of the draft model’s input. 
Furthermore, EAGLE-3 \cite{li2025eagle} adopts a training-time test technique to avoid relying solely on the top layer and shifts from feature prediction to direct token prediction.
Overall, self-speculative decoding enables more efficient draft prediction with less training data and smaller model size. Therefore, our work focuses on this paradigm to further enhance decoding efficiency in long-context generation.

\paragraph{Efficient Long-Context Generation}
The main bottlenecks in long-context generation lie in the rapidly growing KV cache and the complexity of attention. To address these challenges, several research directions have emerged.
KV cache quantization \cite{liu2024kivi, hooper2024kvquant} reduces memory usage and accelerates attention through faster data loading.
Sparse attention \cite{streamingllm, tang2024quest, zhang2023h2o} leverages the inherent sparsity of attention patterns, allowing models to compute attention over only a small subset of key–value pairs with minimal accuracy loss. 
New architectures, such as Linear Attention and Hybrid Linear Attention \cite{team2025kimi}, have also been actively explored as alternatives to full attention with superior efficiency.

Triforce \cite{suntriforce} introduces streaming cache drafting and proposes a hierarchical speculation to alleviate memory bottlenecks. MagicDec \cite{sadhukhanmagicdec} leverages a draft model with sparse KV cache to accelerate generation for large batch size. TokenSwift \cite{wu2025tokenswift} employs partial KV cache and a Medusa-style draft model to achieve lossless acceleration for ultra long sequence generation.
The aforementioned methods primarily focus on the draft phase, whereas our work targets the verification phase, adopting an EAGLE-3–style draft model and partial verification to further accelerate long-context generation.

\section{SpecPV}
\label{section: specpv}

In this section, we introduce SpecPV, a partial verification self-speculative decoding approach for long-context generation.
We first introduce the inference framework of self-speculative decoding and identify its bottlenecks.
Then, we present our partial verification method.
Finally, to minimize potential performance degradation, we periodically adopt full verification to mitigate accumulated errors. 
The following subsections describe each component in detail.

\subsection{Self-Speculative Decoding Framework}
\label{section: self-spec}

\begin{figure}[t]
\begin{center}
\centerline{\includegraphics[width=\columnwidth]{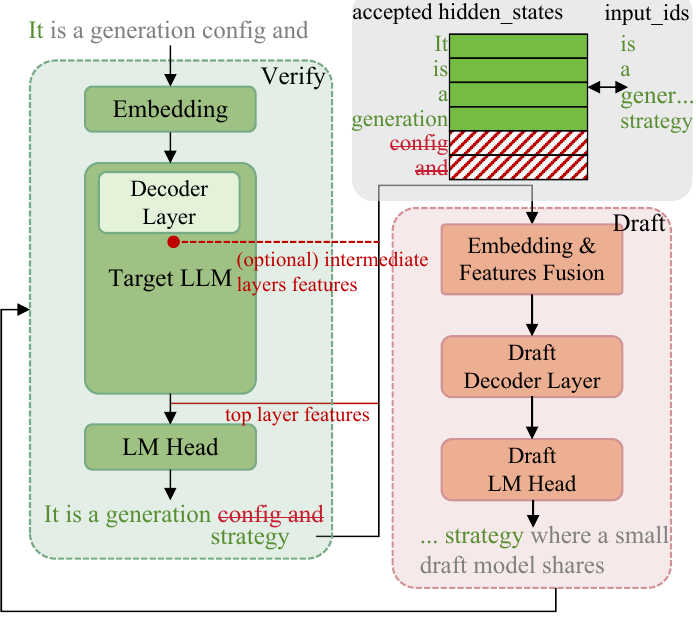}}
\vskip 0.1in
\caption{Illustration of the generation process in self-speculative decoding. A key characteristic of self-speculation is that the draft model reuses the layer features from the target LLM.}
\label{fig: self-spec framework}
\end{center}
\vspace{-2.5em}
\end{figure}

Traditional approaches typically employ a standalone small speculative model that shares the target LLM’s vocabulary. In practice, this model is usually a smaller member of the same model family. In contrast, the key characteristic of self-speculative decoding is that the draft model reuses layer features from the target LLM. We provide an example in Figure \ref{fig: self-spec framework} to better illustrate it.

After the prefilling stage, the model predicts the next token and generates a set of layer features that serve as inputs for self-speculative drafting. Different methods make use of these features in distinct ways: some rely solely on top-layer features, while others combine token representations with layer features.
These methods also differ in how they produce multiple future tokens. Multi-head designs assign each head to a specific future position, whereas single-head approaches repeatedly apply one prediction head in an autoregressive manner. 
The draft module itself is typically lightweight, often consisting of only a single decoder layer.

A common misunderstanding is that drafting requires running the target LLM to obtain layer features, making it appear as though the target model must participate in the draft phase. In practice, verification already produces all necessary layer features for the accepted tokens. The features are naturally aligned with the next step predictions, in which the feature corresponding to token $x_i$ is used to draft token $x_{i+1}$.
As a result, the draft phase consists solely of the draft module computation, without invoking any additional forward passes of the target LLM.

In long-context scenarios, most prior works primarily explore alternative drafting strategies while still relying on full verification to guarantee lossless acceleration. However, as shown in Section \ref{sec:Introduction}, we observe that as the context length grows, the verification stage progressively becomes the dominant performance bottleneck. This motivates our work to focus on accelerating the verification stage, aiming to substantially improve the efficiency of speculative decoding in long-context generation.

\subsection{Partial Verification}
\label{section: partial}
As illustrated in Figure \ref{fig: specpv framework}, we adopt partial verification to improve the decoding efficiency. Since current KV cache managers commonly organize the cache in block formats \cite{kwon2023efficient}, we also select partial KV cache at the block level.

The partial KV cache consists of four components.
First, due to the \emph{attention sink} mechanism \cite{streamingllm}, keeping the KV blocks of the initial tokens will largely preserve the performance of window attention. Therefore, we always retain the first few KV cache blocks as \emph{sink tokens}.
Next is the \emph{retrieval tokens}, which forms the main body of the partial KV cache. Inspired by Quest \cite{tang2024quest}, the set of critical tokens depends on the current query. 
Owing to the semantic continuity of natural language, which endows attention with intrinsic locality \cite{yang2025lserve}, the partial KV cache does not need to be updated at every step while still maintaining strong model performance. To further accelerate the retrieval process, we cache the key-states summary of each block.
\begin{equation}
C_i = \left( K_i^{\max},\; K_i^{\min} \right)
\end{equation}
where $K_i^{\max}$ and $K_i^{\min}$ denote the element-wise maximum and minimum of the key states within block \emph{i}.
Next, we compute a score $s_i$ for each block based on the input query states.
During standard decoding, only a single token is generated at each step, and thus only one query state is produced. However, in the verification phase, the model receives a sequence of candidate tokens generated by the draft module, resulting in multiple query states. To accommodate this situation, we first compute per-query scores for each block and then aggregate them into a single importance score.
\begin{align}
s_{i,j} &= \max\!\left( q_j \left(K_i^{\max}\right)^{\top},\; q_j \left(K_i^{\min}\right)^{\top} \right) \label{eq:block-score} \\
s_i &= f\!\left( s_{i,1}, s_{i,2}, \dots, s_{i,M} \right) \label{eq:block-score-reduced}
\end{align}
where \(s_{i,j}\) denotes the score between the \(j\)-th query state and the key state of block \(i\), and \(f\) is a reduction function (e.g., max, mean, or last), with \emph{last} referring to the query state of the most recently verified token.

Additionally, we retain a small number of recent \emph{local tokens}, maintained as a fixed-size window.
The last component is a buffer. It keeps tokens that have been partially verified and are awaiting correction. During forward phase, the key states of all candidate tokens are temporarily kept in this buffer so that invalid tokens can be removed once verification is completed. 
Although we describe these four components separately, they are contiguous in token order and are handled as a single continuous KV segment during the attention forward pass.

\subsection{Rectified with Full Verification}
\label{section: rectify}
During decoding, the errors introduced by partial verification gradually accumulate, causing the generated output to drift over time. In practice, this accumulation comes from two sources: the key–value states of partially verified tokens are inaccurate, and the retrieved context changes as generation progresses. A practical solution is to periodically apply full verification to eliminate these accumulated errors.

Before entering the verification forward pass, the partially verified tokens are prepended to the newly generated candidate tokens. During verification, we perform the attention computation using the full KV cache and refresh the partial KV cache accordingly.
In this forward pass, we compute key–value states for both the partially verified tokens and the candidate tokens.
Afterward, we update the retrieval tokens and the local window tokens. Finally, once the candidate tokens have been evaluated, we remove the invalid tokens from both the partial and the full KV cache.

We illustrate the three verification processes in Figure \ref{fig: specpv framework}. 
When the sequence length is smaller than the partial KV cache budget, partial verification is disabled. As the sequence grows and exceeds the partial budget, we perform a full verification once to initialize the partial KV cache, then smoothly transition into partial verification. 
Additionally, we set a maximum buffer size for the partial KV cache.
When the total number of partially verified and candidate tokens exceeds this buffer, we switch back to full verification and refresh the partial KV cache. By adjusting the buffer size, we can control how frequently full verification is triggered.

Furthermore, when memory is constrained, a common strategy for long-context inference is KV cache offloading.
Since the partial and draft cache are significantly smaller than the full KV cache, we offload only the full KV cache to host memory. At each layer, we keep only its key and value states on device and prefetch the next layer when needed. Under this setting, partial verification is especially helpful because it reduces full cache access and lowers PCIe transfer overhead, thus improving inference efficiency. Overall, the framework of SpecPV is summarized in Appendix \ref{app: specpv}.

\section{Experiments}
\label{section: experiments}
In this section, we design a series of experiments to evaluate SpecPV in both model performance and generation efficiency.

\subsection{Setup}
\paragraph{Models}
We evaluate SpecPV on LLaMA-3.1-8B-Instruct \cite{grattafiori2024llama} and the Qwen-3 series (4B, 8B and 14B) \cite{yang2025qwen3}. 
For the draft module, we adopt the EAGLE-3 framework \cite{li2025eagle}, a state-of-the-art approach for self-speculative decoding.
The publicly released EAGLE-3 models are trained with a 2K context window. Building on these released weights, we further apply YARN-based long-context adaptation \cite{peng2023yarn} using 6,400 PG-19 samples at a 32K sequence length.
Additional details are provided in Appendix \ref{app: yarn}.

\paragraph{Tasks}
For efficiency evaluation, we construct a story continuation task on PG-19 \cite{pg19}, which allows us to measure speedup under different context lengths. 
For quality evaluation, we use tasks from LongBench v1 \cite{bai2024longbench} and v2 \cite{bai2024longbench2}, covering two categories: document summarization and question answering.

\paragraph{Baselines}
We compare the efficiency of SpecPV against three speculative decoding baselines:
\begin{itemize}[itemsep=1pt, topsep=1pt, parsep=1pt]
    \item \textbf{TriForce:} TriForce \cite{suntriforce} employs an independent draft model and performs hierarchical speculation for long-context generation. The original method supports only LLaMA-2 models, and pretraining a standalone draft model from scratch would require substantial resources. Therefore, for LLaMA-3, we use Qwama-0.5B-Instruct \cite{qwama0.5b} as the draft model, which is based on Qwen2-0.5B-Instruct but adapted to the LLaMA-3 vocabulary.
    \item \textbf{TokenSwift:} TokenSwift \cite{wu2025tokenswift} adopts a Medusa-style draft head and uses a partial KV cache for drafting. It primarily focuses on ultra–long sequence generation, and therefore incorporates additional techniques such as token reutilization and contextual penalty.
    \item \textbf{EAGLE3-YARN:} We directly use our YARN-adapted EAGLE-3 models for speculative decoding generation.
\end{itemize}

\begin{table*}[t]
\caption{Experimental results of LLaMA3.1-8B-Instruct across different context length. We compare SpecPV with various baselines using partial KV caches of 8K, 4K, and 2K. In the table, $\alpha$ denotes the micro-averaged throughput speedup, while $\tau$ represents the macro-averaged draft accept length.}
\label{tab:speedup llama3}
\begin{center}
\begin{small}
\begin{tabular}{c cccccc}
\toprule
\multirow{2}{*}{\textbf{Method}} &
\textbf{10K} & \textbf{20K} & \textbf{30K} &
\textbf{40K} & \textbf{50K} & \textbf{60K} \\
\cmidrule(lr){2-7}
& $\alpha$ \qquad $\tau$ &
  $\alpha$ \qquad $\tau$ &
  $\alpha$ \qquad $\tau$ &
  $\alpha$ \qquad $\tau$ &
  $\alpha$ \qquad $\tau$ &
  $\alpha$ \qquad $\tau$ \\
\midrule
TriForce
& 1.72$\times$ \; 2.62
& 2.44$\times$ \; 2.82
& 2.65$\times$ \; 2.61
& 3.10$\times$ \; 2.80
& 3.42$\times$ \; 2.82
& 3.17$\times$ \; 2.61 \\
TokenSwift
& 1.41$\times$ \; 1.64
& 1.70$\times$ \; 1.73
& 1.83$\times$ \; 1.73
& 1.98$\times$ \; 1.77
& 2.06$\times$ \; 1.77
& 2.11$\times$ \; 1.75 \\
EAGLE3-YaRN
& 2.49$\times$ \; 3.38
& 2.74$\times$ \; 3.49
& 2.82$\times$ \; 3.38
& 3.02$\times$ \; 3.57
& 3.06$\times$ \; 3.46
& 3.11$\times$ \; 3.48 \\
SpecPV-8K
& 2.53$\times$ \; 3.34
& 3.54$\times$ \; 3.51
& 4.23$\times$ \; 3.42
& 4.88$\times$ \; 3.56
& 5.31$\times$ \; 3.46
& 5.80$\times$ \; 3.57 \\
SpecPV-4K
& 2.92$\times$ \; 3.29
& 4.00$\times$ \; 3.57
& 4.57$\times$ \; 3.43
& 5.46$\times$ \; 3.67
& 5.66$\times$ \; 3.52
& 6.15$\times$ \; 3.56 \\
SpecPV-2K
& 2.88$\times$ \; 3.30
& 3.83$\times$ \; 3.42
& 4.75$\times$ \; 3.46
& 5.21$\times$ \; 3.55
& 5.79$\times$ \; 3.56
& 6.29$\times$ \; 3.57 \\
\bottomrule
\end{tabular}
\end{small}
\end{center}
\vskip -0.1in
\end{table*}

\paragraph{Metrics}
We evaluate both the generation quality and the efficiency of SpecPV.
For efficiency, we report two metrics:
\begin{itemize}[itemsep=1pt, topsep=1pt, parsep=1pt]
    \item \textbf{Overall Speedup $\alpha$:} The decoding time of all methods is compared against the standard autoregressive baseline.
Speedup is computed as:
\begin{equation}
\text{Speedup} = \frac{\text{Throughput}_{\text{SpecPV}}}{\text{Throughput}_{\text{AR}}}
\end{equation}
    \item \textbf{Draft Accept Length $\tau$:} We measure the average number of tokens accepted per verification step. Our metric allows an accept length of 0 when all drafted tokens are rejected.
\end{itemize}
For quality evaluation, we adopt task-appropriate metrics:
\begin{itemize}[itemsep=1pt, topsep=1pt, parsep=1pt]
    \item \textbf{Question Answering:} We measure exact-match accuracy, comparing the generated answer with the ground truth after normalization.
    \item \textbf{Summarization:} We measure ROUGE-L \cite{lin2004rouge}, which measures longest common subsequence overlap,
 and BLEURT \cite{sellam2020bleurt}, a learned metric that evaluates semantic similarity between sentences.
\end{itemize}

\begin{figure}[t]
\begin{center}
\centerline{\includegraphics[width=\columnwidth]{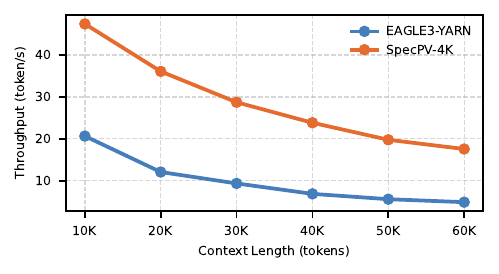}}
\caption{Decoding throughput of LLaMA3.1-8B-Instruct on a single RTX 4090 GPU with KV cache offloading. }
\label{fig: offload}
\end{center}
\vspace{-3em}
\end{figure}

\paragraph{Implementations}
For consistency across baselines, we implement all attention operations using the scaled dot product attention of Pytorch. We evaluate decoding efficiency on an A100 80GB GPU with context lengths up to 60K. In memory-constrained settings, we test efficiency on a single RTX 4090 24GB GPU with KV cache offloading.

\subsection{Generation Efficiency}

\begin{table}[t]
\caption{Summarization performance on GovReport and QMSum (temperature = 0) on Qwen3-8B and LLaMA3.1-8B-Instruct. RL denotes ROUGE-L and BLT denotes BLEURT.}
\label{tab: sum_results}
\begin{center}
\begin{small}

\begin{tabular}{c c ccc ccc}
\toprule
\multirow{2}{*}{\textbf{Models}} & \multirow{2}{*}{\textbf{Budget}} &
\multicolumn{2}{c}{\textbf{GovReport}} &
\multicolumn{2}{c}{\textbf{QMSum}} \\
\cmidrule(lr){3-6}
& &
RL & BLT &
RL & BLT \\
\midrule
\multirow{4}{*}{LLaMA3.1}
& Full & 35.3 & 42.7 & 25.0 & 44.9 \\
& 4096 & 34.4 & 42.6 & 25.0 & 44.1 \\
& 2048 & 33.8 & 42.5 & 24.9 & 44.2 \\
& 512  & 30.7 & 42.3 & 24.3 & 44.2 \\
\midrule
\multirow{4}{*}{Qwen3}
& Full & 33.7 & 41.4 & 23.9 & 42.4 \\
& 4096 & 33.3 & 41.3 & 23.6 & 42.0 \\
& 2048 & 33.1 & 41.3 & 23.8 & 41.9 \\
& 512  & 31.1 & 41.3 & 22.8 & 41.2 \\
\bottomrule
\end{tabular}

\end{small}
\end{center}
\vspace{-1.8em}
\end{table}

We construct a test dataset from PG-19 by extracting segments of varying lengths while ensuring that sentences remain as complete as possible. 
Each model is then prompted to continue the story, generating up to 1024 tokens with a temperature of 0.
For SpecPV generation, we use mean reduction for KV cache retrieval and we size the buffer to accommodate all tokens processed in a single verification step with an additional margin of 20. 
We only modify the verification process and keep the drafting process unchanged.

\begin{table*}[t]
\caption{Experimental results on the Qwen3 series across different context length. We report the throughput speedup ($\alpha$) and draft accept length ($\tau$) under various partial KV cache budgets. For the 60K context length, we set the YARN scaling factor of Qwen3 to 2.0 to accommodate the extended context window.}
\label{tab: speedup-qwen}
\small
\vskip -0.2in
\begin{center}
\begin{small}
\begin{tabular}{cccccccc}
\toprule
\multirow{2}{*}{\textbf{Size}} & \multirow{2}{*}{\textbf{Method}} &
\textbf{10K} & \textbf{20K} & \textbf{30K} &
\textbf{40K} & \textbf{50K} & \textbf{60K} \\
\cmidrule(lr){3-8}
& &
$\alpha$ \qquad $\tau$ &
$\alpha$ \qquad $\tau$ &
$\alpha$ \qquad $\tau$ &
$\alpha$ \qquad $\tau$ &
$\alpha$ \qquad $\tau$ &
$\alpha$ \qquad $\tau$ \\
\midrule
\multirow{4}{*}{4B} &
EAGLE3-YaRN
& 2.43$\times$ \; 3.27 & 2.59$\times$ \; 3.28 & 2.61$\times$ \; 3.21
& 2.54$\times$ \; 3.05 & 2.70$\times$ \; 3.13 & 2.79$\times$ \; 3.18 \\
& SpecPV-8K
& 2.46$\times$ \; 3.35 & 3.50$\times$ \; 3.30 & 4.02$\times$ \; 3.23
& 4.50$\times$ \; 3.14 & 5.04$\times$ \; 3.11 & 5.52$\times$ \; 3.11 \\
& SpecPV-4K
& 2.55$\times$ \; 3.37 & 3.40$\times$ \; 3.11 & 4.22$\times$ \; 3.17
& 4.73$\times$ \; 3.17 & 5.44$\times$ \; 3.28 & 5.89$\times$ \; 3.29 \\
& SpecPV-2K
& 2.76$\times$ \; 3.39 & 3.69$\times$ \; 3.31 & 4.31$\times$ \; 3.25
& 4.84$\times$ \; 3.20 & 5.32$\times$ \; 3.15 & 5.82$\times$ \; 3.14 \\
\midrule
\multirow{4}{*}{8B} &
EAGLE3-YaRN
& 2.44$\times$ \; 3.23 & 2.65$\times$ \; 3.25 & 2.79$\times$ \; 3.35
& 2.78$\times$ \; 3.35 & 2.78$\times$ \; 3.29 & 2.83$\times$ \; 3.22 \\
& SpecPV-8K
& 2.39$\times$ \; 3.02 & 3.33$\times$ \; 3.19 & 4.00$\times$ \; 3.19
& 4.46$\times$ \; 3.25 & 4.97$\times$ \; 3.27 & 5.31$\times$ \; 3.11 \\
& SpecPV-4K
& 2.44$\times$ \; 2.98 & 3.47$\times$ \; 3.06 & 4.13$\times$ \; 3.02
& 4.83$\times$ \; 3.37 & 5.21$\times$ \; 3.29 & 5.44$\times$ \; 3.06 \\
& SpecPV-2K
& 2.53$\times$ \; 2.90 & 3.58$\times$ \; 3.05 & 4.14$\times$ \; 3.01
& 4.61$\times$ \; 3.13 & 5.24$\times$ \; 3.32 & 5.35$\times$ \; 3.01 \\
\midrule
\multirow{4}{*}{14B} &
EAGLE3-YaRN
& 2.46$\times$ \; 3.03 & 2.74$\times$ \; 3.30 & 2.83$\times$ \; 3.34
& 3.12$\times$ \; 3.81 & 3.18$\times$ \; 3.78 & 2.98$\times$ \; 3.31 \\
& SpecPV-8K
& 2.46$\times$ \; 2.91 & 3.35$\times$ \; 3.11 & 3.83$\times$ \; 2.98
& 5.06$\times$ \; 3.79 & 5.45$\times$ \; 3.57 & 5.63$\times$ \; 3.44 \\
& SpecPV-4K
& 2.75$\times$ \; 2.89 & 3.68$\times$ \; 3.14 & 4.13$\times$ \; 3.02
& 5.37$\times$ \; 3.71 & 6.02$\times$ \; 3.69 & 5.83$\times$ \; 3.23 \\
& SpecPV-2K
& 2.74$\times$ \; 2.79 & 3.73$\times$ \; 3.08 & 4.37$\times$ \; 3.10
& 5.37$\times$ \; 3.66 & 5.99$\times$ \; 3.60 & 5.73$\times$ \; 2.96 \\
\bottomrule

\end{tabular}
\end{small}
\end{center}
\vskip -0.1in
\end{table*}
\begin{figure*}[t]
\begin{center}
\centerline{\includegraphics[width=1.9\columnwidth]{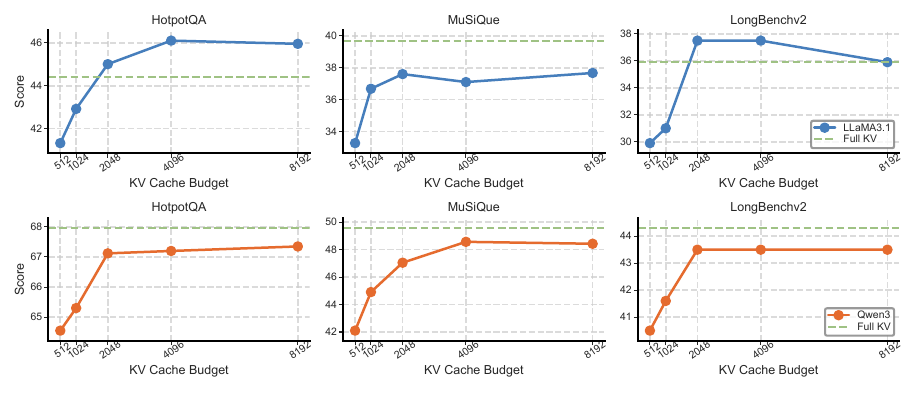}}
\caption{Accuracy on QA tasks under different partial KV cache budgets. For LongBench v2, samples exceeding 64K context length are excluded. For all datasets, we first generate a chain of thought and then prompt the model to produce a standardized final answer. For most datasets, SpecPV achieves performance comparable to full verification under a 4096 token KV budget.}
\label{fig: longbench acc}
\end{center}
\vskip -0.3in
\end{figure*} 

We compare SpecPV with other baselines on LLaMA3.1-8B-Instruct across different context length. The results are shown in Table \ref{tab:speedup llama3}. The partial KV budget for both TriForce and TokenSwift is set to 4K.
For TokenSwift, its design primarily targets ultra-long sequence generation, typically beyond 20K tokens. Under our evaluation settings, techniques such as token reutilization have limited impact, resulting in relatively modest speedup. 
For TriForce, the hierarchical verification design yields solid acceleration in long-context settings. However, it does not refresh the partial KV cache, and each verification step relies on full verification to remain lossless. In addition, its standalone draft model yields slightly shorter draft accepted length in our experiments. These factors together contribute to its slower decoding speed compared with SpecPV.
After applying our YARN-based long-context adaptation, the EAGLE-3 model also achieves notable draft accept length in long-context scenarios.
When the context is relatively short, the speedup from partial verification is limited. However, as the context further increases, the benefits become far more significant. We also observe that different partial KV budgets lead to variations in draft accept length. In practice, the partial KV budget should be chosen flexibly based on the actual context length in order to balance efficiency and accuracy.
Furthermore, we also evaluate performance on a single RTX 4090 GPU with KV cache offloading, shown in Figure \ref{fig: offload}. Since the primary bottleneck in this setting is PCIe transfer, and the partial KV cache is small enough to avoid host and device transfers, SpecPV achieves significant speedup.

We also evaluate SpecPV under different partial KV cache budgets on the Qwen3 series. For continuation tasks, we use the no-think mode. The results are shown in Table \ref{tab: speedup-qwen}.
Due to its mixed think and no-think output modes, Qwen3 produces slightly shorter draft accept length than LLaMA3.1.
Overall, SpecPV provides consistent acceleration across different LLMs and model size.

\subsection{Generation Quality}
We evaluate generation quality on both summarization and question answering tasks. 
For summarization, we use the GovReport \cite{huang2021efficient} and QMSum \cite{zhong2021qmsum} datasets from LongBenchv1. 
Following standard practice, we measure the similarity between model-generated summaries and the human-provided ground truth using ROUGE-L and BLEURT. The results are shown in Table~\ref{tab: sum_results}. Compared to full decoding, SpecPV achieves comparable end-task performance across different cache budgets, with only minor performance degradation. These results indicate that SpecPV largely preserves the semantic quality of the generated summaries while enabling more efficient inference.

For question answering, we use HotpotQA \cite{yang2018hotpotqa} and MuSiQue \cite{trivedi2022musique} from LongBenchv1 together with LongBenchv2 samples after excluding those that exceed a 64K context length.
We first generate a chain of thought (CoT) and then prompt the model to produce a standardized final answer. 
For Qwen3, we generate the CoT in think mode to obtain stronger performance. 
The results are shown in Figure \ref{fig: longbench acc}.
Overall, as the KV cache budget decreases, model accuracy gradually declines. For most datasets, SpecPV achieves performance comparable to full verification with a 4096 token KV budget. 
In addition, we hypothesize that discarding part of the KV cache can sometimes reduce long-context noise, which explains why SpecPV even outperforms full verification on certain datasets. 
Overall, SpecPV maintains generation quality that is highly consistent with full verification.

Additionally, we conduct ablation studies on SpecPV hyperparameters, including buffer size and reduction type, in Appendix \ref{app: ablation}. And provide detailed generation settings in Appendix \ref{app: params}.


\subsection{Case Study}
\begin{figure}[t]
\begin{center}
\centerline{\includegraphics[width=\columnwidth]{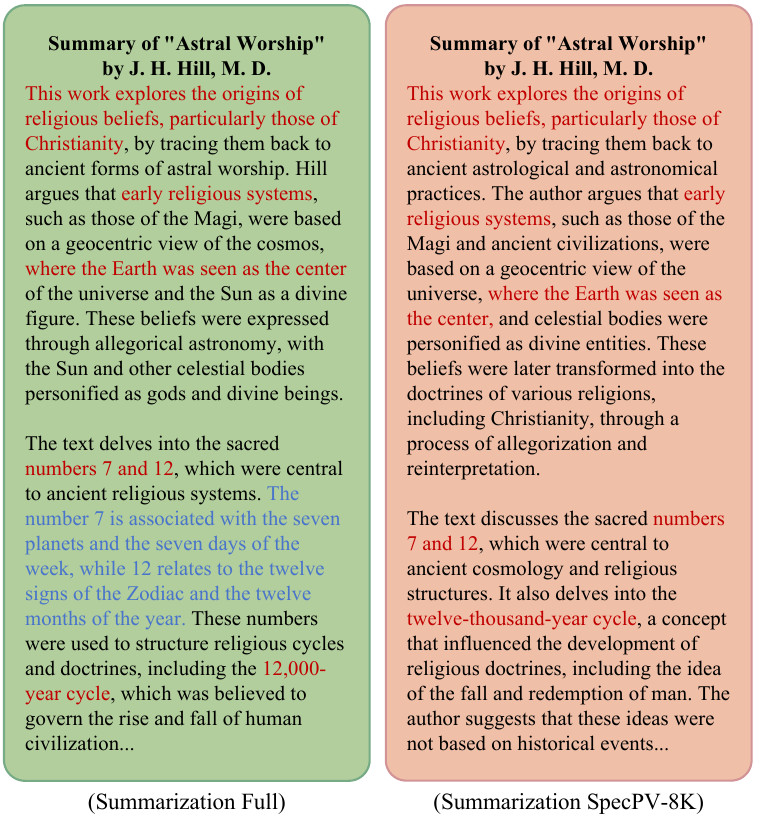}}
\vskip 0.05in
\caption{A case study on Qwen3-8B, where the LLM summarizes a 50K tokens document using full verification and SpecPV.}
\label{fig: case}
\end{center}
\vskip -0.35in
\end{figure}

Figure \ref{fig: case} presents a case study on Qwen3-8B for book summarization. We compare the output generated by full verification and SpecPV. 
From this example, we observe that although SpecPV uses partial KV verification, it preserves most of the key information, and the crucial factual content (highlighted in red) remains correct. 
However, the output generated with full verification includes more fine-grained details, shown in blue. 
Because SpecPV relies on only a subset of the KV cache, we conjecture that certain fine-grained details may be lost, which is not well reflected by standard accuracy metrics.
We hope future long-context generation methods will better address this limitation.

\section{Conclusion}
In this work, we propose SpecPV, a self-speculative decoding approach that performs fast verification using partial KV cache and periodically applies full verification to eliminate accumulated errors. By accelerating the verification stage within speculative decoding, SpecPV further improves generation efficiency in long-context settings. Our experiments show that SpecPV achieves up to 6× decoding speedup compared with standard autoregressive decoding, with only minor accuracy degradation. We believe this work contributes to enabling more efficient and practical speculative decoding for long-context LLM applications.

\section*{Limitations}
While SpecPV demonstrates strong efficiency and quality trade-offs, several aspects merit further study. Compared to standard lossless speculative decoding methods, SpecPV introduces a controlled approximation via partial KV verification, as shown in Figure~\ref{fig: case}, it preserves key semantic and factual content but may occasionally miss fine-grained details, which are not always reflected by standard metrics. In addition, its acceleration performance depends on the draft model (here, an EAGLE3-style self-speculative decoding), and may vary across languages or application scenarios. Finally, recent advances in pre-training attention architectures further improve native long-context modeling; exploring how speculative decoding can be combined with such approaches is a promising direction for future work.



\bibliography{custom}

\appendix
\section{SpecPV Algorithm Description}
\label{app: specpv}

We provide the pseudocode of SpecPV in Algorithm \ref{alg: specpv} for clarity.

\section{YARN-Based Extension of the EAGLE-3 Module}
\label{app: yarn}

For the original EAGLE-3 draft module, directly applying it to long-context scenarios fails to produce effective draft predictions. The core issue is that the decoder layer inside EAGLE-3 relies on RoPE position encoding, which does not extrapolate well to extended context lengths. Moreover, most publicly available EAGLE-3 models are trained on self-generated conversational data \cite{ding2023enhancing} with context windows around 2K tokens. 
As a result, we first need to extend its positional embeddings.

We extend the context length using YARN. Since YARN finetuning only repairs the positional embedding and does not require injecting additional knowledge, it can be trained with a relatively small amount of data. We construct a dataset of 6,400 samples from PG-19 with a 32K context window.
We fine-tune the EAGLE-3 module based on SpecForge \cite{specforge2025}, and our training hyperparameters are listed in Table \ref{tab: eagle3yarn}.

\begin{algorithm}[t]
   \caption{SpecPV Generation}
   \label{alg: specpv}
 \small
\begin{algorithmic}[1]
\STATE \textbf{Input:} target model $\mathcal{M}_{\mathrm{tgt}}$, draft model $\mathcal{M}_{\mathrm{draft}}$, 
prompt $p$, cache configuration $\mathcal{C}$, sampling configuration $\mathcal{S}$,
max length $\mathcal{T}$, max new tokens $\mathcal{N}$
\STATE \textbf{Output:} generated sequence $y$

\STATE $\mathrm{KV}_{\mathrm{full}}, \mathrm{KV}_{\mathrm{partial}}, \mathrm{KV}_{\mathrm{draft}} \gets \text{InitKV}(\mathcal{C})$
\STATE $h,x \gets \text{ChunkPrefill}(\mathcal{M}_{\mathrm{tgt}}, KV_{\mathrm{full}}, \mathcal{M}_{\mathrm{draft}}, KV_{\mathrm{draft}}, p)$
\STATE $y \gets \text{concat}(p, [x])$

\FOR{$t = 1$ \textbf{to} $\mathcal{T}$}
    \STATE $\text{candidates} \gets \text{TreeDraft}(\mathcal{M}_{\mathrm{draft}}, \mathrm{KV}_{\mathrm{draft}}, h, x)$
	\STATE $\text{mode} \gets \text{SelectMode}(\text{len}(y),\; KV_{\mathrm{partial}})$

	\IF{$\text{mode} = \textsc{Full}$}
    	\STATE $KV_{\mathrm{in}} \gets KV_{\mathrm{full}}$
	\ELSIF{$\text{mode} = \textsc{Partial}$}
    	\STATE $KV_{\mathrm{in}} \gets KV_{\mathrm{partial}}$
	\ELSIF{$\text{mode} = \textsc{Refresh}$}
    	\STATE $KV_{\mathrm{in}} \gets \{KV_{\mathrm{partial}},\, KV_{\mathrm{full}}\}$
    	    	\STATE $\text{candidates} \gets \text{concat}(x_{\mathrm{partial}}, \text{candidates})$
	\ENDIF

	\STATE $\text{logits},\; h \gets \text{TreeDecode}(\mathcal{M}_{\mathrm{tgt}},\; \text{candidates},\; KV_{\mathrm{in}})$
    
    \STATE $x \gets \text{PostEvaluate}(\text{candidates}, \text{logits}, \mathrm{KV}_{\mathrm{in}}, \mathcal{S})$
    \STATE Append $x$ to $y$
	
	\IF{$\text{len}(y)-\text{len}(p) \ge \mathcal{N} \OR \textsc{EOS} \in x$}
    	\STATE \textbf{break}
	\ENDIF

\ENDFOR

\end{algorithmic}
\end{algorithm}

\begin{table}[h]
\caption{Training hyperparameters for fine-tuning the EAGLE-3 module using SpecForge.}
\label{tab: eagle3yarn}
\small
\begin{center}
\begin{small}
\begin{tabular}{l c}
\toprule
\textbf{Hyperparameter} & \textbf{Value} \\
\midrule
Num epochs                       & 1 \\
Learning rate                    & 2e-5 \\
Warmup ratio                     & 0.05 \\
Max length                       & 32768 \\
Training-TimeTest length                       & 4 \\
Tensor parallel size (TP)        & 1 \\
Data parallel size (DP)          & 4 \\
Draft micro batch size           & 1 \\
Draft global batch size          & 16 \\
Draft accumulation steps         & 4 \\
Scaling factor & 16.0\\
\bottomrule
\end{tabular}
\end{small}
\end{center}
\vspace{-1em}
\end{table}

After training, we set the scaling factor to 32.0, which allows the model to extrapolate to a 64K context length directly.
To enable the EAGLE-3 module to perform autoregressive generation more effectively, the authors adopt the training-time test technique. During training, the model performs multiple rounds of generation, allowing the loss to be aggregated across multiple decoding steps:
\begin{equation}
L=L_0+\beta L_1+\beta^2 L_2 +\cdots+\beta ^nL_n
\end{equation}

\begin{figure*}[h]
\vskip 0.1in
\begin{center}
\centerline{\includegraphics[width=2\columnwidth]{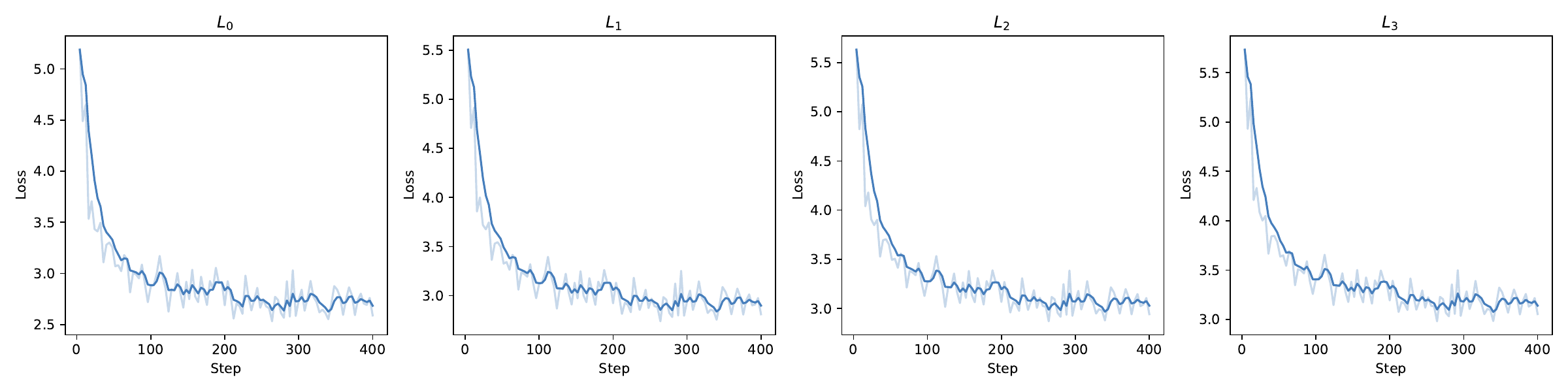}}
\caption{Loss curves of EAGLE3 YARN fine-tuning on LLaMA3.1-8B-Instruct. The plots show both the raw loss and its time-weighted exponential moving average (EMA) across training steps.}
\label{fig: loss_curve}
\vspace{-1em}
\end{center}
\end{figure*}

Following common practice, we set the discount factor $\beta$ to 0.8.
The training loss curves are shown in Figure \ref{fig: loss_curve}. As the training-time test step increases, the loss consistently decreases.

\section{Ablation Study}
\label{app: ablation}

In this appendix, we focus on the impact of different query-reduction strategies and the influence of buffer size.

We evaluate different reduction strategies on Qwen3-8B and LLaMA-3.1-8B-Instruct using the GovReport dataset under a 2K KV budget. 
To more precisely characterize the effects of different reduction types and refresh intervals, we directly compare the model-generated summarization outputs with those obtained under full verification.
The results are shown in Table \ref{tab:query-reduction}. Overall, the choice of reduction type has only a minor effect on performance. Mean reduction yields the highest similarity and also produces slightly longer draft accept length. Consequently, we adopt mean reduction in all other experiments.

\begin{figure}[t]
\begin{center}
\centerline{\includegraphics[width=\columnwidth]{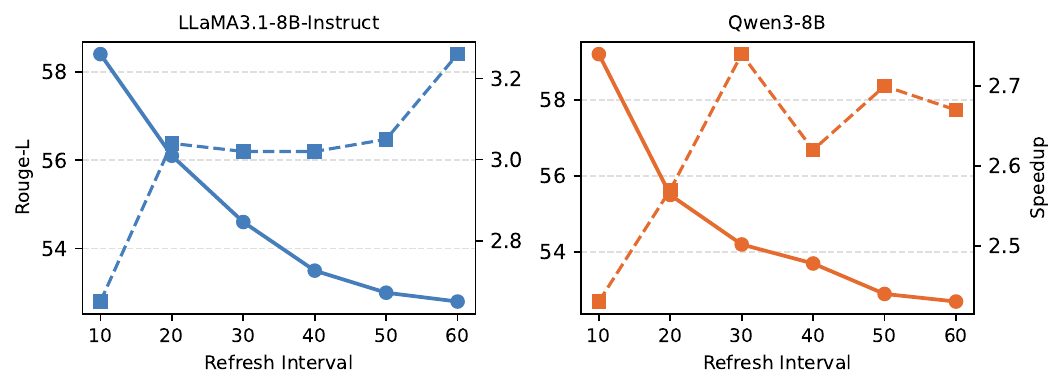}}
\vskip 0.05in
\caption{Experimental results of different refresh intervals on QMSum. Solid lines indicate ROUGE-L, and dashed lines indicate speedup. Overall, larger refresh intervals lead to greater deviation from full verification, but yield higher speedup.}
\label{fig: ablation}
\end{center}
\vskip -0.3in
\end{figure}

By adjusting the buffer size, we effectively control the interval for partial KV cache refreshes. We evaluate this effect on QMSum, as shown in Figure \ref{fig: ablation}. 
Across different models, we observe a similar trend: as the refresh interval increases, the similarity between SpecPV and full verification gradually decreases, indicating that periodically partial KV cache updates help preserve model performance. 
To balance efficiency and accuracy, we typically set the buffer size to the number of tokens involved in one verification step plus 20.

\begin{table}[t]
\caption{Different reduction results on GovReport. Mean, Max, and Last denote different retrieval score aggregation strategies.}
\label{tab:query-reduction}
\small
\begin{center}
\setlength{\tabcolsep}{3pt}  
\begin{small}

\begin{tabular}{c ccc ccc ccc}
\toprule
\multirow{2}{*}{\textbf{Models}} &
\multicolumn{2}{c}{\textbf{Mean}} &
\multicolumn{2}{c}{\textbf{Max}} &
\multicolumn{2}{c}{\textbf{Last}} \\
\cmidrule(lr){2-7}
& RL & $\tau$ &
  RL & $\tau$ &
  RL & $\tau$  \\
\midrule
LLaMA3.1
& 56.1 & 3.53
& 55.8 & 3.52
& 54.0 & 3.47 \\
Qwen3
& 56.0 & 2.77
& 55.5 & 2.76
& 55.4 & 2.77 \\
\bottomrule
\end{tabular}

\end{small}
\end{center}
\vskip -0.1in
\end{table}

\section{Generation Settings}
\label{app: params}

For the LLaMA 3.1-8B-Instruct model, which supports up to a 128K context length and operates only in a instant response mode, we set the generation temperature to 0 across all experiments.

For the Qwen3 series, which provides both thinking and non-thinking modes with officially recommended configurations, we adopt different settings depending on the task. In the thinking mode, we use temperature = 0.6, TopP = 0.95, and TopK = 20. In the non-thinking mode, we set temperature = 0. For summarization and continuation tasks, we directly use the non-thinking mode, while for question answering tasks, we use the thinking mode.

In addition, for the SpecPV algorithm, we set the block size to 16, the number of sink tokens to 32, and the local window size to 128 tokens. The buffer size is fixed at 20.

\section{Ablation on Temperature}

\begin{table}[t]
\small
\centering
\caption{Summarization performance under different temperatures and partial KV budgets. RL denotes ROUGE-L and BLT denotes BLEURT.}
\label{tab:temp_summary}
\begin{tabular}{c c cc cc}
\toprule
\multirow{2}{*}{\textbf{Temperature}} & \multirow{2}{*}{\textbf{Budget}} 
& \multicolumn{2}{c}{\textbf{GovReport}} 
& \multicolumn{2}{c}{\textbf{QMSum}} \\
\cmidrule(lr){3-4} \cmidrule(lr){5-6}
& & \textbf{RL} & \textbf{BLT} & \textbf{RL} & \textbf{BLT} \\
\midrule

\multirow{4}{*}{$t=0.7$}
& Full & 34.8 & 42.4 & 24.9 & 43.9 \\
& 4096 & 34.5 & 42.3 & 24.5 & 43.7 \\
& 2048 & 33.4 & 42.5 & 24.4 & 43.6 \\
& 512  & 31.7 & 42.1 & 23.9 & 43.0 \\
\midrule

\multirow{4}{*}{$t=1.0$}
& Full & 34.5 & 42.2 & 24.2 & 44.0 \\
& 4096 & 33.9 & 42.1 & 23.9 & 43.3 \\
& 2048 & 32.9 & 42.4 & 24.2 & 43.1 \\
& 512  & 30.0 & 41.8 & 23.4 & 43.2 \\
\bottomrule
\end{tabular}
\end{table}
\begin{table*}[t]
\caption{Experimental results under different context lengths and temperatures for LLaMA3.1-8B-Instruct. 
$\alpha$ denotes the throughput speedup over autoregressive decoding, 
and $\tau$ represents the average draft accept length.}
\label{tab:appendix_speedup}
\begin{center}
\begin{small}
\begin{tabular}{c c cccccc}
\toprule
\multirow{2}{*}{\textbf{Temp.}} &
\multirow{2}{*}{\textbf{Method}} &
\textbf{10K} & \textbf{20K} & \textbf{30K} &
\textbf{40K} & \textbf{50K} & \textbf{60K} \\
\cmidrule(lr){3-8}
& &
$\alpha$ \qquad $\tau$ &
$\alpha$ \qquad $\tau$ &
$\alpha$ \qquad $\tau$ &
$\alpha$ \qquad $\tau$ &
$\alpha$ \qquad $\tau$ &
$\alpha$ \qquad $\tau$ \\
\midrule

\multirow{3}{*}{$t=0.7$}

& EAGLE3-YARN
& 2.22$\times$ \; 3.12
& 2.53$\times$ \; 3.18
& 2.65$\times$ \; 3.13
& 2.81$\times$ \; 3.26
& 2.90$\times$ \; 3.31
& 2.88$\times$ \; 3.13 \\

& SpecPV-8K
& 2.31$\times$ \; 3.05
& 3.16$\times$ \; 3.12
& 3.95$\times$ \; 3.17
& 4.45$\times$ \; 3.19
& 4.98$\times$ \; 3.20
& 5.49$\times$ \; 3.30 \\

& SpecPV-4K
& 2.66$\times$ \; 3.02
& 3.64$\times$ \; 3.14
& 4.16$\times$ \; 3.17
& 4.92$\times$ \; 3.29
& 5.49$\times$ \; 3.24
& 5.77$\times$ \; 3.24 \\

\midrule

\multirow{3}{*}{$t=1.0$}

& EAGLE3-YARN
& 2.20$\times$ \; 2.70
& 2.25$\times$ \; 2.61
& 2.42$\times$ \; 2.68
& 2.52$\times$ \; 2.75
& 2.65$\times$ \; 2.81
& 2.67$\times$ \; 2.81 \\

& SpecPV-8K
& 2.06$\times$ \; 2.42
& 2.73$\times$ \; 2.46
& 3.50$\times$ \; 2.55
& 4.20$\times$ \; 2.70
& 4.65$\times$ \; 2.64
& 5.16$\times$ \; 2.81 \\

& SpecPV-4K
& 2.45$\times$ \; 2.58
& 3.14$\times$ \; 2.66
& 3.96$\times$ \; 2.67
& 4.54$\times$ \; 2.76
& 5.18$\times$ \; 2.78
& 5.46$\times$ \; 2.85 \\

\bottomrule
\end{tabular}
\end{small}
\end{center}
\vskip -0.1in
\end{table*}
For non-deterministic sampling, we evaluate the performance of LLaMA3.1-8B-Instruct under different temperature settings. We fix $top_k=0$ and $top_p=0.9$, following commonly adopted sampling configurations in LLM generation. Specifically, we conduct experiments with $T=0.7$ and $T=1.0$, and analyze both the summarization quality and acceleration performance under non-deterministic decoding.

We observe that different temperature settings have relatively limited impact on the overall generation quality. SpecPV maintains comparable ROUGE-L and BLEURT scores to full verification under non-deterministic decoding, while a small KV budget (e.g., 512) leads to a more noticeable ROUGE-L drop due to the loss of long-context information, although the semantic quality remains largely preserved. In terms of acceleration, SpecPV shows consistent speedup trends across different temperatures. However, the average accept length becomes smaller when $T=1.0$, since higher temperatures produce smoother token distributions and increase the discrepancy between the draft and target model distributions, resulting in more frequent draft-target mismatches during speculative decoding.

\end{document}